\documentclass[a4paper]{article}
\pdfoutput=1 
\usepackage{amsmath,amssymb,amsfonts}
\usepackage{subcaption}
\usepackage{tabularx}
\usepackage[colorlinks]{hyperref}
\usepackage{graphicx}
\begin{document}

\title{Stroke extraction for offline handwritten mathematical expression recognition \thanks{Declarations of interest: none}}
\author{Chungkwong Chan\thanks{School of Mathematics, Sun Yat-Sen University, 135 Xingang Xi Road, Guangzhou, 510275, China} \thanks{Email address: chsongg@mail2.sysu.edu.cn or chan@chungkwong.cc}}

\maketitle

\begin{abstract}
	Offline handwritten mathematical expression recognition is often considered much harder than its online counterpart due to the absence of temporal information. In order to take advantage of the more mature methods for online recognition and save resources, an oversegmentation approach is proposed to recover strokes from textual bitmap images automatically. The proposed algorithm first breaks down the skeleton of a binarized image into junctions and segments, then segments are merged to form strokes, finally stroke order is normalized by using recursive projection and topological sort. Good offline accuracy was obtained in combination with ordinary online recognizers, which are not specially designed for extracted strokes. Given a ready-made state-of-the-art online handwritten mathematical expression recognizer, the proposed procedure correctly recognized 58.22\%, 65.65\%, and 65.22\% of the offline formulas rendered from the datasets of the Competitions on Recognition of Online Handwritten Mathematical Expressions(CROHME) in 2014, 2016, and 2019 respectively. Furthermore, given a trainable online recognition system, retraining it with extracted strokes resulted in an offline recognizer with the same level of accuracy. On the other hand, the speed of the entire pipeline was fast enough to facilitate on-device recognition on mobile phones with limited resources. To conclude, stroke extraction provides an attractive way to build optical character recognition software.
\end{abstract}

\begin{description}
	\item[Keywords] Character recognition; feature extraction; offline handwritten mathematical expression recognition; optical character recognition software; stroke extraction
	\item[2010 MSC] 68T10; 68T45; 68U10
	
\end{description}

\section{Introduction}
\label{intro}

Mathematical expressions appear frequently in engineering and scientific documents. Since they contain valuable information, digitizing them would maximize their usability by enabling retrieval\cite{Zanibbi2012Recognition}, integration to semantic web\cite{Marchiori2003}, and other automated tasks. Compared with ordinary text, mathematical expressions can present some concepts more concisely because of their two-dimensional structure. However, such compact representations are difficult to be recognized mechanically.

People are used to writing mathematical expressions on paper or blackboard, so it is inconvenience to input them by using another way. Traditional input devices like keyboards are designed for sequence of characters, although spatial relations between symbols can be represented by markups such as TeX or MathML, entering mathematical expressions by typing in a computer language is not user-friendly at all, as new users are asked to learn a new language, remember a lot of commands and deal with miscellaneous errors. On the other hand, entering mathematical expressions with a graphical equation editor by choosing structural elements and symbols from toolboxes is inefficient for frequent users.

A recognition system turns handwritten mathematical expressions into machine manipulable syntax trees. Online recognition enables people to take notes or solve equations by writing on a touch-based device or dragging a mouse. On the other hand, offline recognition enables people to digitize existing manuscripts by scanning or record lecture notes on blackboards by taking photos. The main difference between two kinds of recognition is that trajectories are available to an online recognizer, whereas only bitmap images are given to an offline recognizer. 

It is not surprising that online handwriting recognition systems often achieve a much higher accuracy. In fact, an online recognition problem can be converted to the corresponding offline problem by simply rendering the strokes. In the opposite direction, if the sequence of strokes can be recovered from a bitmap image, an online recognizer can also be applied to do offline recognition\cite{NISHIDA19951213}. The objective of this paper is to explore the feasibility of offline handwriting recognition via stroke extraction without designing specialized online recognizers.

The main concern is that recovery of strokes is unlikely to be perfect, so it may become a single point of failure. Fortunately, good online recognition systems can tolerate some input errors, since diversity of writing habits already affected them. Furthermore, if the underlying online recognition system has been retrained with extracted strokes, it should be able to adapt to them. In this case, accuracy of stroke extraction needs not be critical to that of offline recognition.

The main contributions of this paper include:
\begin{enumerate}
	\item We propose a stroke extraction algorithm for handwritten mathematical expression, which enables reduction from offline recognition to online recognition.
	\item We verify that the accuracy of the proposed system is comparable to that of other recent offline handwritten mathematical expression recognition systems, given a ready-made state-of-the-art online recognizer. 
	\item We demonstrate that the gap in accuracy between the proposed offline system and the underlying online system can be narrowed significantly by retraining.
\end{enumerate}

The remainder of this paper is divided into four sections. Section \ref{sec:related} reviews methods to recognize online and offline mathematical expressions, as well as methods to recover trajectories. Section \ref{sec:algorithm} describes the proposed stroke extraction algorithm for mathematical expression in detail. Section \ref{sec:evaluation} presents experimental results on standard datasets. Section \ref{sec:conclusion} concludes the paper.

\section{Related works}
\label{sec:related}

\subsection{Online handwritten mathematical expression recognition}
\label{subsec:relatedOnline}

Online handwritten mathematical expression recognition is a long-standing problem, a lot of works have been done since Anderson\cite{Anderson1967Syntax}. In the past decade, the problem attracted more and more attention because of the Competitions on Recognition of Online Handwritten Mathematical Expressions(CROHME)\cite{Mouch2016Advancing}, which were held in 2011, 2012, 2013, 2014, 2016, and 2019.

Traditionally, the problem is further divided into symbol recognition and structural analysis\cite{Zanibbi2012Recognition}. For example, \'Alvaro et al.\cite{ALVARO201458} applied hidden Markov model to recognize the symbols and parser for a predefined two-dimensional stochastic context free grammar to analyze the structure.

In order to disambiguate symbols with structural information and vice versa, several approaches were proposed to integrate symbol recognition and structural analysis closely. Yamamoto et al.\cite{yamamoto} suggested parsing handwritten mathematical expression directly from strokes by using the Cocke-Younger-Kasami algorithm, so that symbol segmentation, character recognition, and structural analysis could be optimized simultaneously. Awal et al.\cite{AWAL201468} introduced another global approach that applies a segmentation hypothesis generator to deal with delayed strokes.

Recently, with the advances in computational power, end-to-end trainable recurrent neural networks became popular because of its ability to learn complex relations. For example, Zhang et al.\cite{8373726} developed an encoder-decoder framework for online handwritten mathematical expression recognition named Track, Attend, and Parse(TAP), which employs a guided hybrid attention mechanism.

\subsection{Offline handwritten mathematical expression recognition}
\label{subsec:relatedOffline}

On the contrary, dedicated work on offline handwritten mathematical expression recognition was almost blank in literature until very recently. An offline task was first added to CROHME in 2019\cite{CROHME2019}. 

In the past, the closest problem addressed was the more constrained problem of printed mathematical expression recognition\cite{Garain2007OCR}. Again, in a typical system, symbols are first segmented and recognized, then the structure of the expression is analyzed\cite{Chan2000}. For instance, in the state-of-the-art system developed by Suzuki et al. \cite{Suzuki2003INFTY}, symbols are extracted by connected component analysis and then recognized by a nearest neighbor classifier, after that, structural analysis is performed by finding a minimum spanning tree in a directed graph representing spatial relations between symbols.

Recently, Deng et al. \cite{pmlr-v70-deng17a} and Zhang et al. \cite{ZHANG2017196,8546031} developed end-to-end trainable neural encoder-decoder models to translate image of mathematical expression into TeX code directly. They are quite similar to models for online recognition, except that convolutional neural networks are prepended to extract features. It should be noted that this method is so general that it can be applied to any image-to-markup problem, grammar of neither mathematical expression nor TeX is required by those systems explicitly because they can be learned from data. 

\subsection{Stroke extraction}
\label{subsec:relatedStroke}

Stroke extraction was studied for offline signature verification\cite{156588} and East Asian character recognition\cite{LIU20012339}. A typical stroke extractor detects candidates of sub-strokes first and then reassembles them into strokes by resolving ambiguities. Sub-strokes can be detected by breaking down the skeleton or approximating the image with geometrical primitives such as polygonal chains.

Lee et al.\cite{156588} designed a set of heuristic rules to trace the skeleton. Boccignone et al.\cite{BOCCIGNONE1993409} tried to reconstruct strokes by joining the pair of adjoining sub-strokes having the smallest difference in direction, length, and width repeatedly. Doermann et al.\cite{Doermann1995} proposed a general framework to integrate various temporal clues.

J\"{a}ger\cite{546812} reconstructed strokes by minimizing total change in angle between successive segments within a stroke. Lau et al.\cite{Lau2002Stroke} selected another cost function taking distance between successive segments and directions of the segments into account. Unfortunately, this kind of formulations is essentially traveling salesman problem which is NP-complete, so computing the optimum efficiently may not be possible if there are more than a few sub-strokes.

In order to prevent explosion of combinations, Kato et al.\cite{877517} restricted themselves to single-stroke script subjecting to certain assumptions on junctions, so that strokes can be extracted by traversal of graph. Nagoya et al.\cite{6424482} extended the technique to multi-stroke script under assumptions on how strokes are intersected.

Nevertheless, the effectiveness of stroke extraction to offline recognition has remained largely untested. Evaluations of existing stroke extraction methods were often performed visually or quantitatively with their only metrics on small private datasets. Resulting accuracy of offline recognition was seldom reported and limited to single character recognition, where the challenge of symbol segmentation was not addressed. Moreover, they strongly rely on specially designed structural matching methods, which can tolerate a variety of deformations, so they may not work well with ordinary methods for online recognition.

\section{Offline to online reduction}
\label{sec:algorithm}

\subsection{Overview}
\label{subsec:overview}

Given a bitmap image containing a mathematical expression, it must be converted to a sequence of strokes before being passed to an online handwritten mathematical expression recognition engine. In more detail, key steps of the proposed offline handwritten mathematical expression recognition system are:

\begin{enumerate}
	\item Adaptive binarization. Convert the possibly colored input image into a black and white image.
	\item Stroke width transformation. Estimate stroke width for each foreground pixel.
	\item Thinning. Skeletonize the binary image.
	\item Decomposition of the skeleton. Break it down into segments and junctions.
	\item Construction of an attributed graph. Segments and junctions form edges and vertexes of the graph respectively.
	\item Simplification of the attributed graph. Remove vertexes and edges which are likely noises from the graph.
	\item Reconstruction of strokes. Merge segments into strokes by a bottom up clustering.
	\item Fixing double-traced strokes. Reuse some segments to join separated strokes.
	\item Determination of stroke direction. Ensure that the order of points in each stroke conforms to the usual practice.
	\item Stroke order normalization. Sort the strokes by the time they are expected to be written.
	\item Online recognition. Invoke an online handwritten mathematical expression recognition engine to recognize the sequence of strokes extracted.
\end{enumerate} 

\subsection{Preprocessing}
\label{subsec:preprocessing}

Since skeleton roughly preserves the shape of strokes but much simpler, it is easier to trace strokes from the skeleton instead of the full image. Before skeletonization, a colored image should be binarized. The input image is first converted to a grayscale image by averaging the color channels(possibly weighted). Among the large number of binarization methods available, Sauvola's method\cite{Sauvola2000Adaptive} is chosen. Compared with global thresholding techniques such as Otsu's method\cite{4310076}, such a local adaptive approach addressed commonly seen degradations including uneven illumination and random background noises. However, pixels that do not belong to the mathematical expression may still be marked foreground, text next to the expression and grid lines on a notebook for instance. Mathematical expression localization and separation are out of the scope of this paper.

\begin{figure}
	\begin{minipage}[b]{.5\linewidth}
		\centering
		\includegraphics[width=.5\linewidth]{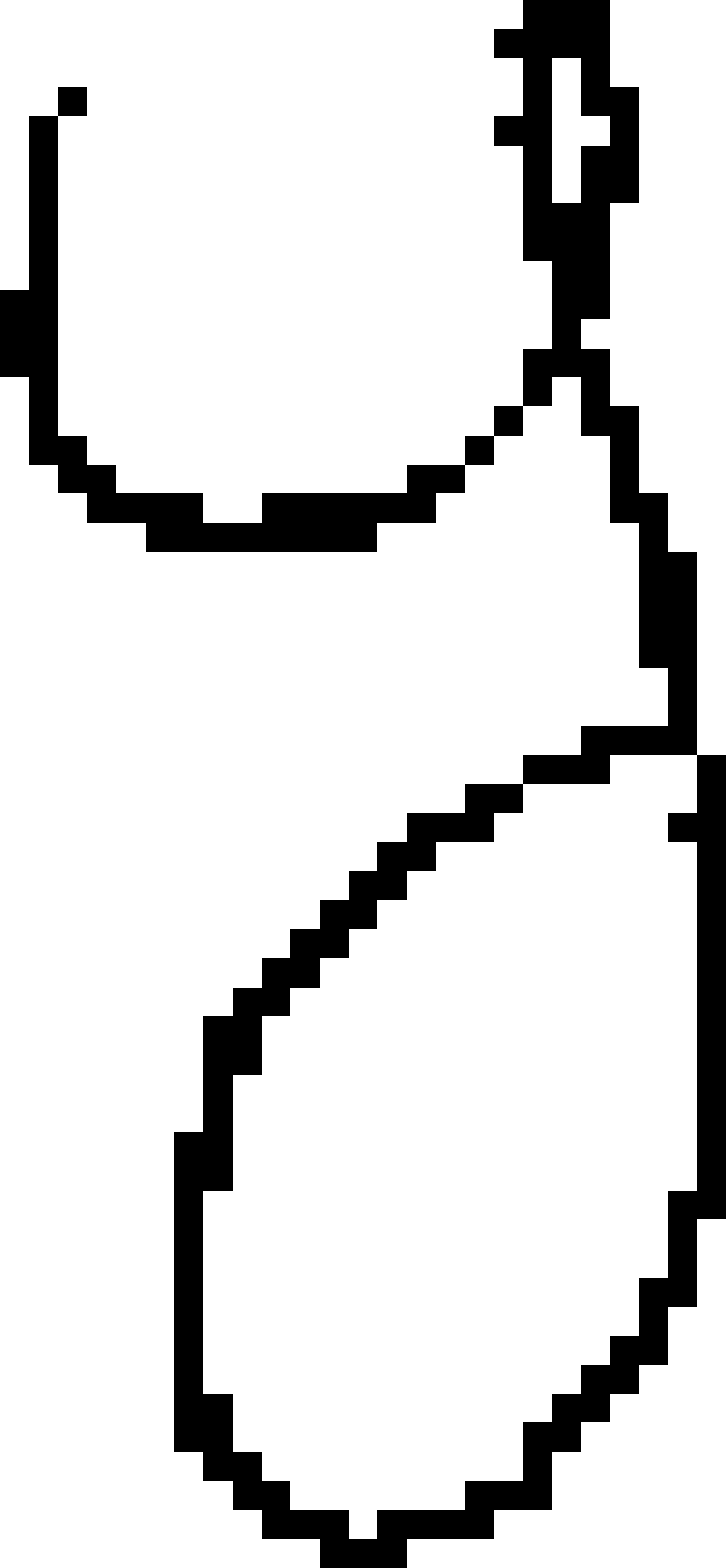}
		\subcaption{Image}\label{fig:beforeThin}
	\end{minipage}%
	\begin{minipage}[b]{.5\linewidth}
		\centering
		\includegraphics[width=.5\linewidth]{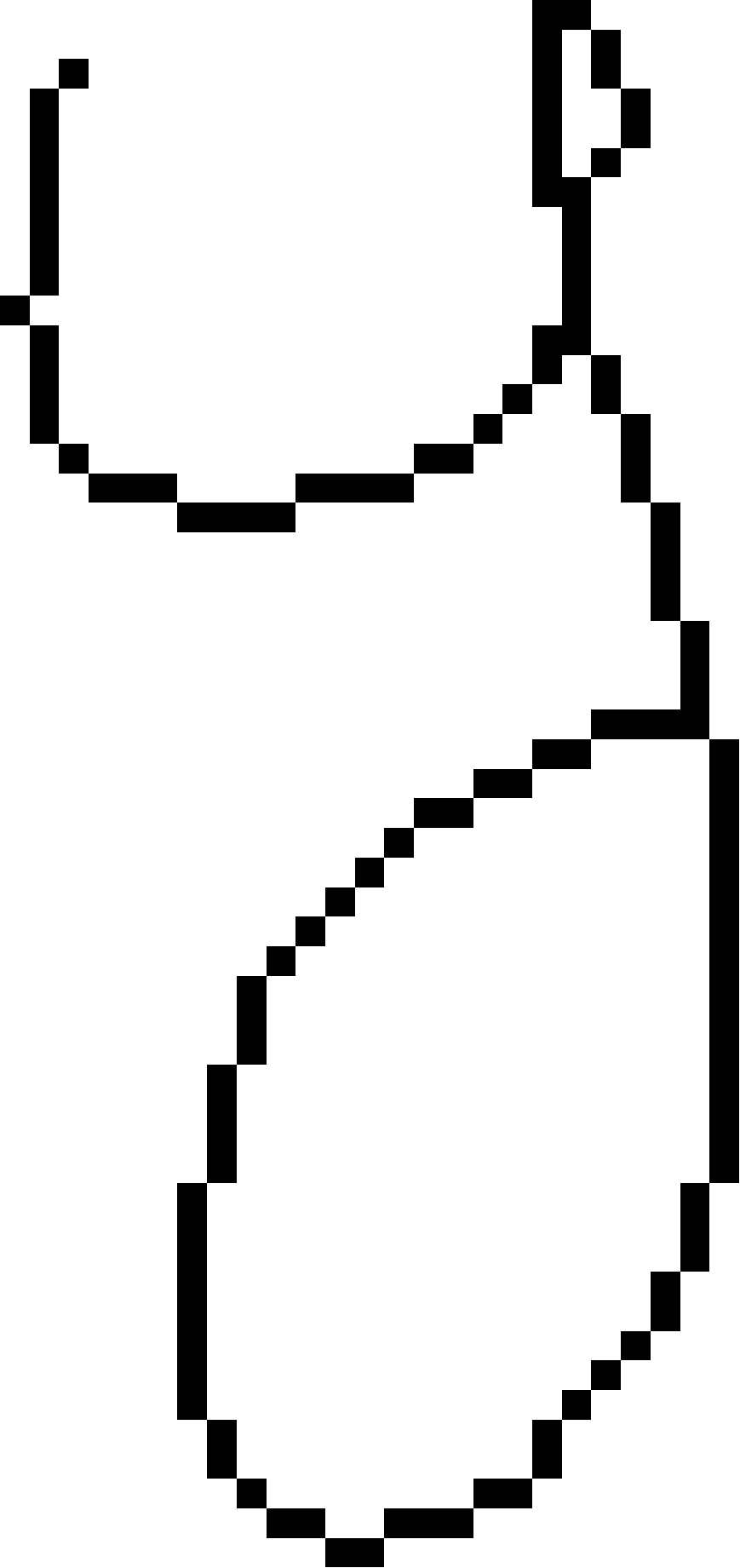}
		\subcaption{Skeleton}\label{fig:afterThin}
	\end{minipage}
	\caption{Thinning}\label{fig:thinning}
\end{figure}

After binarization, skeleton of the image is obtained by using a thinning method by Wang and Zhang \cite{Wang1989A}, which is a variant of the original method by Zhang and Suen\cite{osti5085064} but preserves the shape of diagonal strokes better. Figure \ref{fig:thinning} compares an image with its skeleton.

For printed document recognition, skew detection and correction are often performed. However, they should not be applied to a single mathematical expression because the number of symbols may not be enough to estimate the angle reliably. To make thing worse in the present situation, symbols from a handwritten formula need not stick to a single baseline, so expressions like ``$x^{x^x}$'' may fool skew estimators based on line detection like Hough transformation.

\subsection{Decomposition of skeleton}
\label{subsec:separation}

After skeletonization, the skeleton is decomposed into segments and junctions, thus the skeleton can be viewed as a graph. 

\begin{figure}
	\begin{subfigure}[b]{\linewidth}
		\centering
		\includegraphics[width=\linewidth]{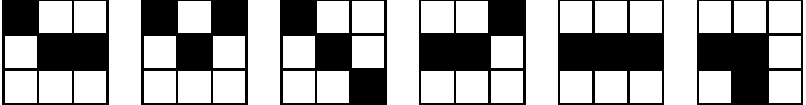}
		\caption{Centers of these $3\times 3$ windows are segment pixels}\label{fig:segmentPixel}
	\end{subfigure}\\
	\begin{subfigure}[b]{\linewidth}
		\centering
		\includegraphics[width=\linewidth]{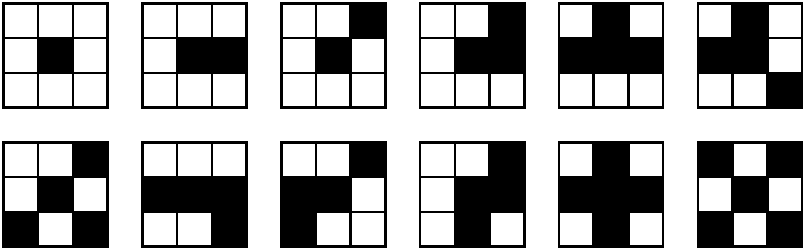}
		\caption{Centers of these $3\times 3$ windows are junction pixels}\label{fig:jointPixel}
	\end{subfigure}
	\caption{Identification of segment pixels and junction pixels}\label{fig:separation}
\end{figure}

A foreground pixel having exactly two other foreground pixels in its 8-neighborhood where the two pixels are not 4-neighbor of each other is called a segment pixel. Other foreground pixels are called junction pixels. Figure \ref{fig:separation} illustrates the rule.

\begin{figure}
	\centering
	\includegraphics[width=0.25\linewidth]{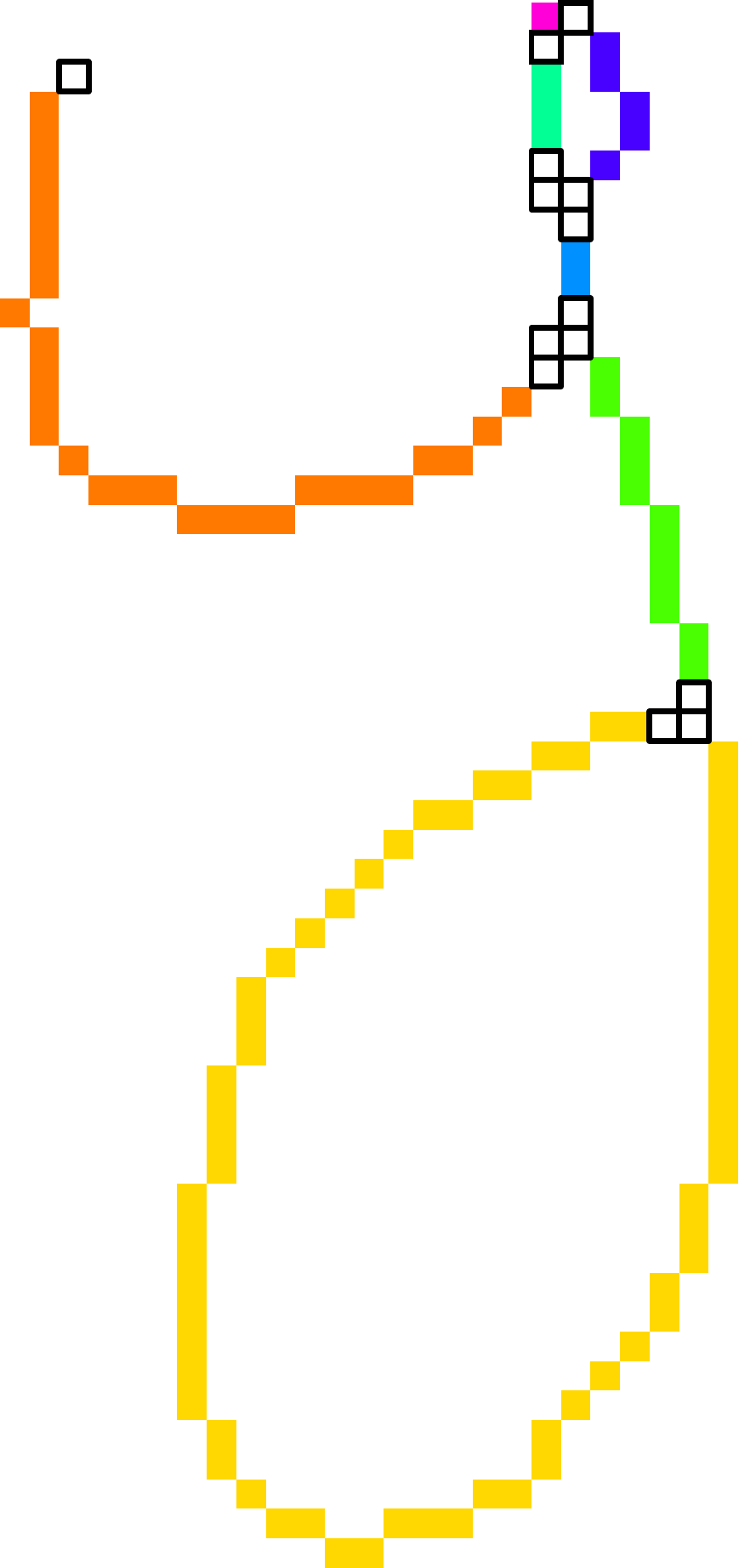}
	\caption{Decomposition of skeleton into segments and junctions}\label{fig:components}
\end{figure}

A connected component of the set of segment pixels is called a segment. On the other hand, a connected component of the set of junction pixels is called a junction. The set of segments and the set of junctions can be computed by using any standard algorithm for connected component analysis\cite{HE201725}. In Figure \ref{fig:components}, segments are solid but junctions are not.

For each segment $S_i$, its pixels can be listed in a way such that successive pixels are 8-neighbor of each other, more formally, $$S_i=\{p_{i,1},\dots,p_{i,\ell_i}\}$$ where $p_{i,k}$ is in the 8-neighborhood of $p_{i,k-1}$ for $k=2,\dots,\ell_i$. If $p_{i,1}$ is in the 8-neighborhood of $p_{i,\ell_i}$, the segment is topologically a circle and does not touch any other junction or segment unless $\ell_i=1$; otherwise, the segment is topologically a line segment, $p_{i,1}$ touches exactly one junction and so do $p_{i,\ell_i}$, other pixels in the segment never touch any other segment or junction.

For the sake of consistency, a ``junction'' is imposed to each looped segment to ensure that every segment has a start pixel and an end pixel, in addition, each touches a junction. Therefore, a junction can be considered as a vertex in the sense of graph theory, whereas a segment can be considered as an edge connecting two (possibly the same) vertexes. Furthermore, a path in this undirected graph corresponds to a possible trace of ink in the input image, a connected component in this graph corresponds to a connected component of the skeleton. Figure \ref{fig:rawGraph} shows the graph coming from the same example as in Figure \ref{fig:components}.

\subsection{Noise reduction}
\label{subsec:noise}

Subtle features such as salt and pepper noises in the binarized image can affect the skeleton, salt noises result in really short segments whereas pepper noises result in isolated junctions. In addition, thinning may introduce distortions. Since they can distract stroke extractor and recognition engine, they should be discarded from the graph. Absolute threshold values are not used because they do not work in all resolutions. Observed that stroke width is uniform in a piece of handwriting, it is chosen to be a reference length.

\begin{figure}
	\centering
	\includegraphics[width=\linewidth]{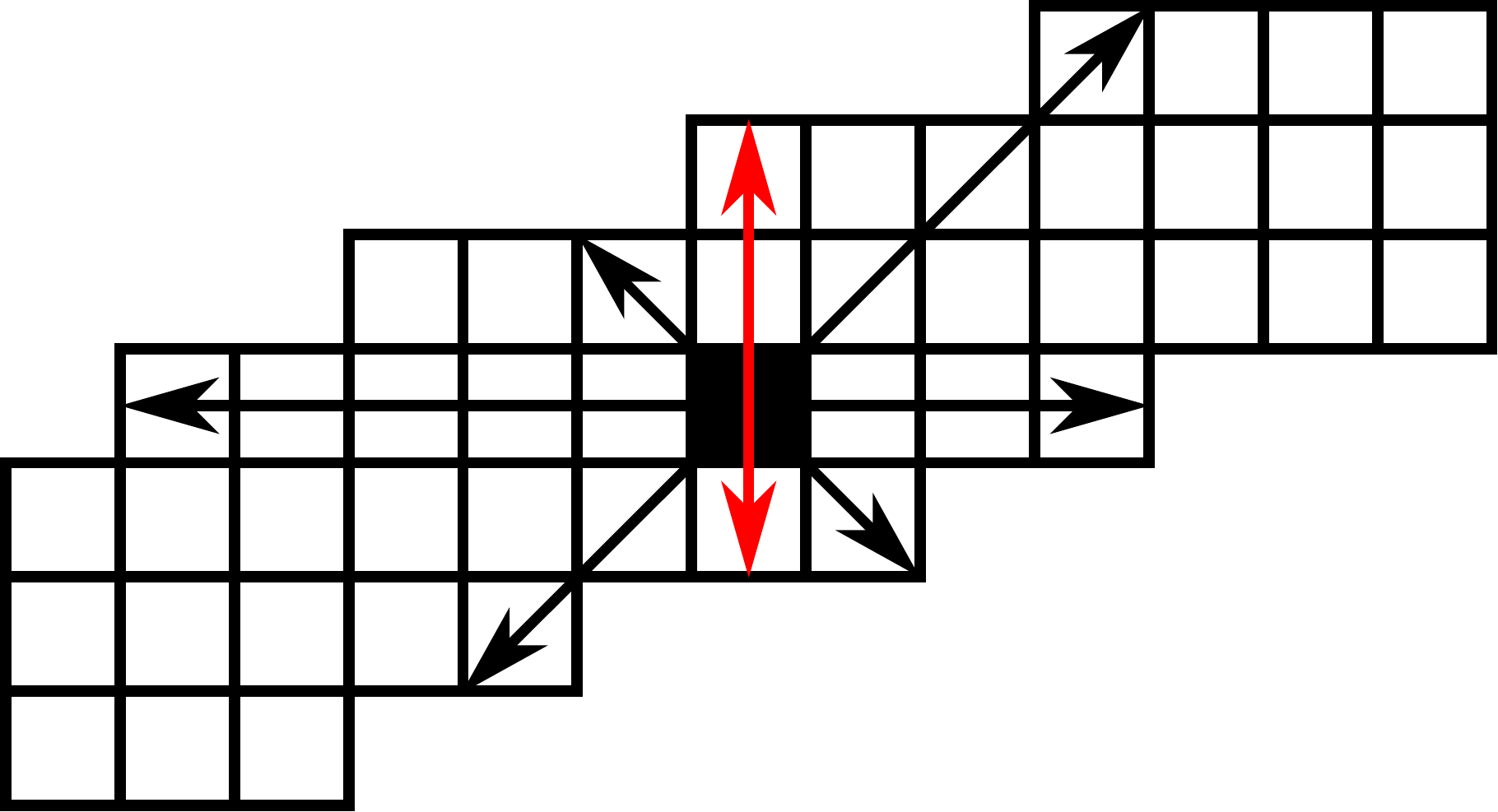}
	\caption{Stroke width is the minimum of four directional run lengths}\label{fig:swt}
\end{figure}

Stroke width transformation is an image operator that assigns an estimated stroke width to each foreground pixel. It was proposed for scene text detection\cite{5540041}, where strokes are considered as contiguous pixels having approximately constant stroke width locally. Using a straightforward viewpoint, stroke width of a pixel can be estimated by the minimum length of the four directional runs\cite{FAN20001881} passing through it as shown in Figure \ref{fig:swt}, where squares represent foreground pixels and arrows represent directional runs of the solid pixel. Under the above definition, stroke width transformation can be computed in linear time with respect to the size of binary image by caching the numbers of successive foreground pixels found in certain directions.

For each set of pixels, its width is estimated by the maximum stroke width among its pixels. Furthermore, the tip size of the pen is estimated by the average stroke width over all the segments. Now, the rules to reduce noises can be stated:

\begin{enumerate}
	\item For each edge with a length smaller than a multiple of the average stroke width, remove it from the graph and merge its end points.
	\item For each vertex with degree 0 and a width less than a multiple of the average stroke width, remove it from the graph.
\end{enumerate}

\begin{figure}
	\begin{minipage}[b]{.45\linewidth}
		\centering
		\includegraphics[width=0.5\linewidth]{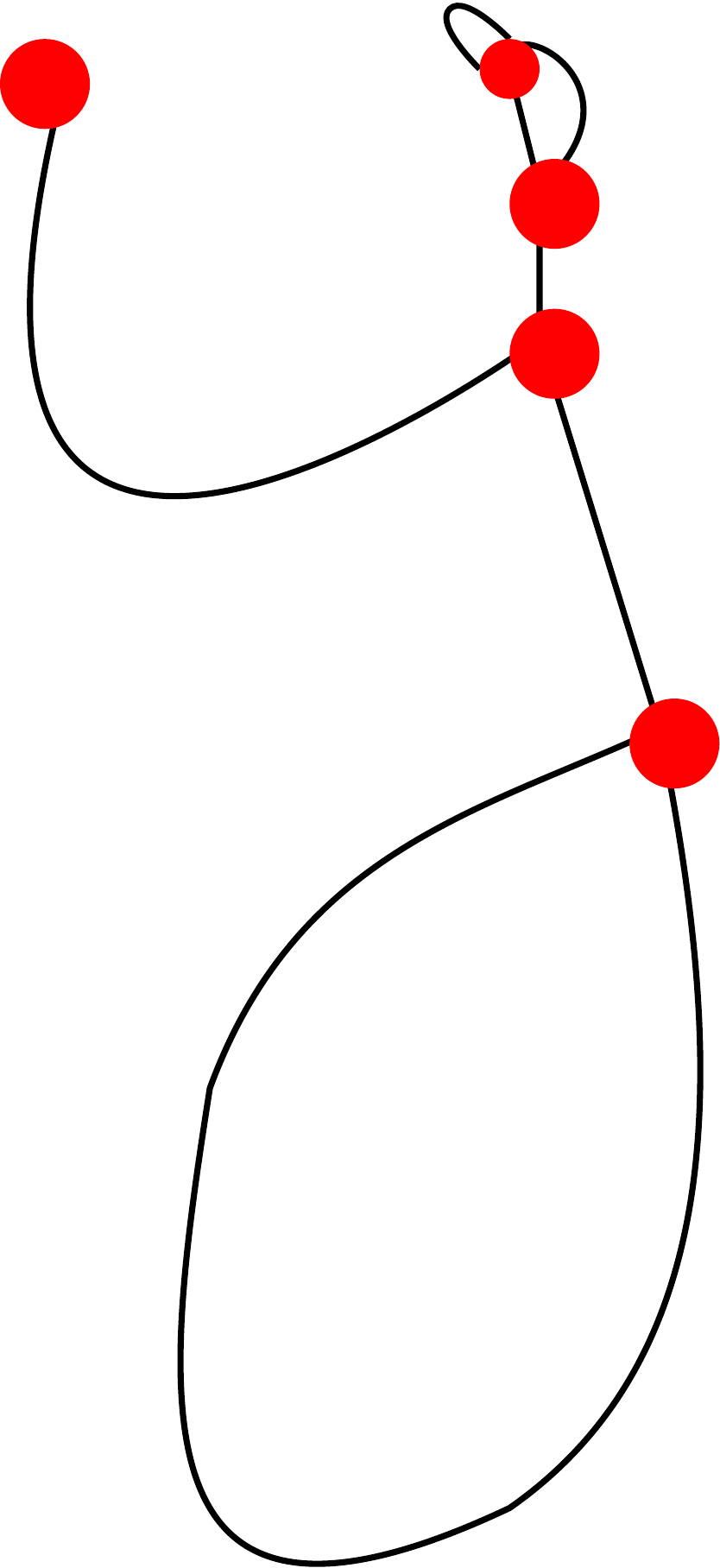}
		\subcaption{Original}\label{fig:rawGraph}
	\end{minipage}
	\begin{minipage}[b]{.45\linewidth}
		\centering
		\includegraphics[width=0.5\linewidth]{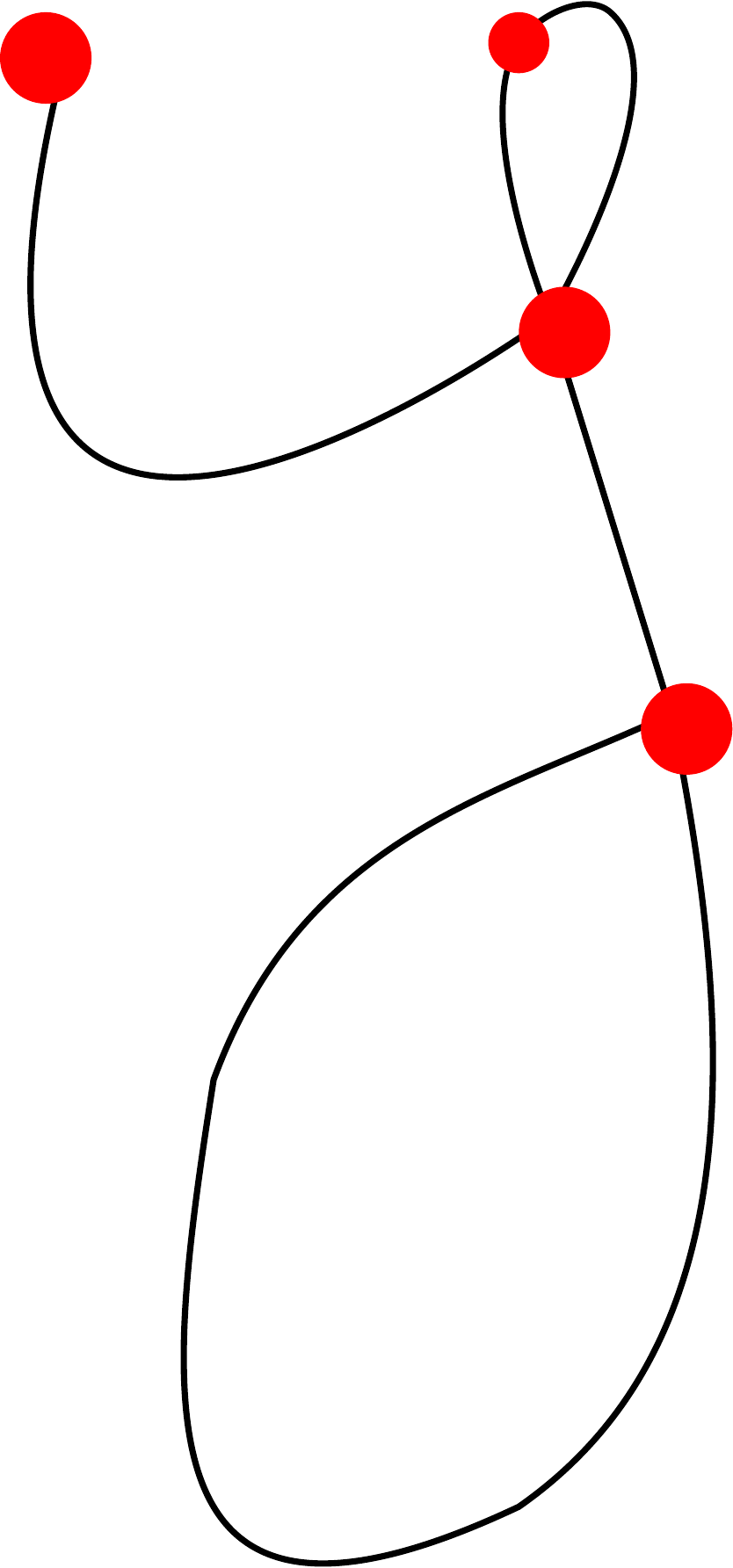}
		\subcaption{Simplified}\label{fig:simplifiedGraph}
	\end{minipage}
	\caption{Skeleton graph}\label{fig:simplify}
\end{figure}

Figure \ref{fig:simplifiedGraph} shows the simplified graph coming from the same example as in Figure \ref{fig:components}.

\subsection{Stroke tracing}
\label{subsec:tracing}

Clearly, an isolated vertex in the graph represents a dot in the mathematical expression, possibly a decimal point or part of a character like ``i''. Therefore, a stroke containing a single point is extracted for each vertex with degree 0. In addition, a path in the skeleton graph indicates a candidate of stroke. Although there may be multiple ways to combine the edges into paths, some combinations are more likely to form strokes of a mathematical expression written by human being. Here are some heuristic principles:

\begin{itemize}
	\item The total number of strokes should be minimal. Since letting the pen to leave the paper requires additional time, an unicursal way is preferred.
	\item The difference in directions between two successive segments should be as small as possible. Since turning suddenly requires slowing down, a fluent stroke is better to be smooth.
\end{itemize}

Subjecting to these considerations, each edge is assigned to exactly one path by a bottom up clustering. Initially, each edge forms a path on its own. While there is a pair of paths having a common end point, choose a pair such that the angle between them is the minimum, then merge them into one path. Repeat the procedure until no path can be merged.

It should be noted that the two principles may not always agree. If the number of strokes is considered more important, its minimum can be obtained by merging each path with circuits that have a common vertex with it, just like the algorithm that search for an Eulerian path.

\subsection{Fixing double-traced strokes}
\label{subsec:double}

\begin{figure}
	\begin{minipage}[b]{.45\linewidth}
		\centering
		\includegraphics[width=\linewidth]{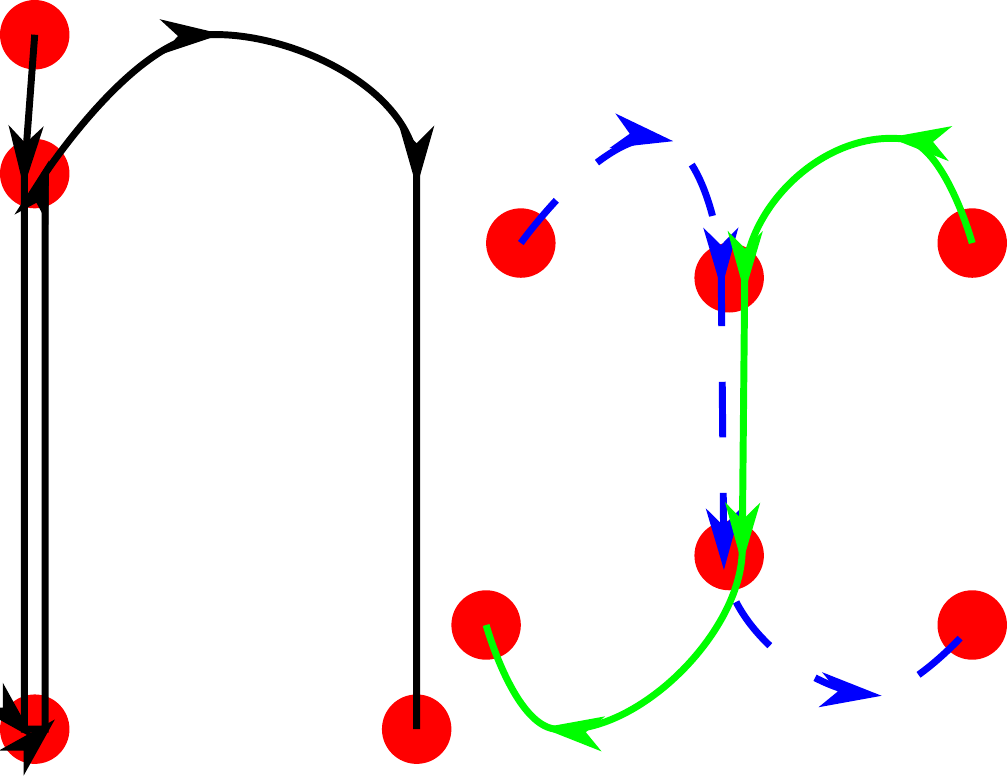}
		\subcaption{Double-traced strokes}\label{fig:strokeMulti}
	\end{minipage}
	\begin{minipage}[b]{.45\linewidth}
		\centering
		\includegraphics[width=\linewidth]{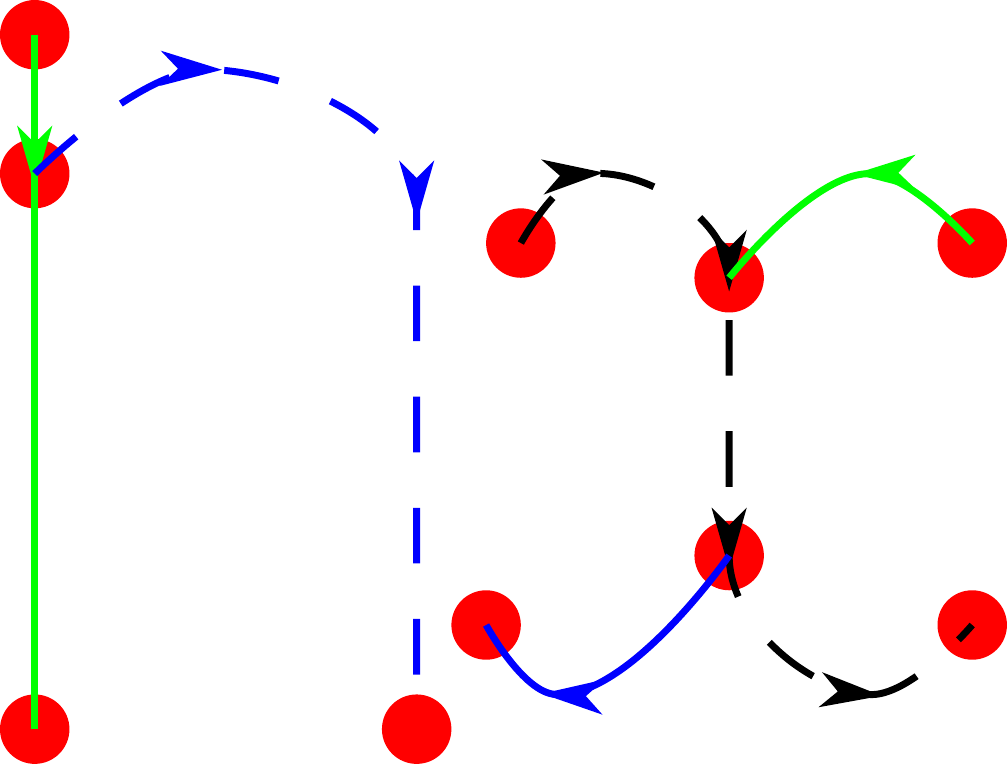}
		\subcaption{Simple strokes}\label{fig:strokeSingle}
	\end{minipage}
	\caption{Identification of double-traced stroke}\label{fig:fix}
\end{figure}

Sometimes, a segment should be shared by more than one strokes or appeared in a stroke multiple times due to reentry during writing as shown in Figure \ref{fig:strokeMulti}. The tracing procedure above would handle such cases incorrectly by producing too many strokes as in Figure \ref{fig:strokeSingle}.

In order to fix the double-traced strokes, a search for shared segments is needed, so that they can be used to reconnect separated strokes. Candidates of shared segments should meet the following criteria:
\begin{itemize}
	\item The segment has two different end points, which are vertexes in the graph with an odd degree. Otherwise, the number of vertexes in the graph having an odd degree does not decrease when it is doubled.
	\item Each end point of the segment is also an end point of a path given by subsection \ref{subsec:tracing} and the angle between them is not close to $\pi/2$. This condition can prevent the two strokes of the symbol ``T'' from being merged.
\end{itemize} 

\subsection{Stroke order normalization}
\label{subsec:ordering}

Firstly, stroke direction detection is performed. The stroke tracing procedure gives rise to an ordering of points within a stroke naturally, however, the opposite ordering may also make sense. Since people usually write from left to right and top to bottom, a simple rule is sufficient to determine the direction of each stroke in most cases. Let the coordinates of the first and the last point of a stroke be $(x_{start},y_{start})$ and $(x_{end},y_{end})$ respectively, then the list of points should be reversed if $$2x_{end}+3y_{end}<2x_{start}+3y_{start}.$$

Finally, the order that people write down the strokes should be recovered. Although some mathematical expression recognition systems are stroke order free\cite{AWAL201468,ALVARO201458}, others use stroke order to prune the search space, so they are sensitive to stroke order\cite{SIMISTIRA201585,Le2016}. In fact, stroke order normalization is a way to turn a stroke order dependent recognizer into a stroke order free one\cite{Le2019}.

There are possibly multiple orders to write down the same formula. For example, someone prefers to write down the square root sign first, but the others write down the radicand first. Therefore, it is not always possible to recover the original order, what can be done is to assign a reasonable order. In some cases, mistakes made by the stroke extractor can also be viewed as a kind of normalization and may in fact enhance performance by eliminating unusual stroke order.

A hierarchical approach is applied to sort the strokes. To begin with, strokes are grouped by recursive projection, then the groups are sorted in a left to right and top to bottom manner. However, recursive X-Y cut cannot determine the order of symbols inside a square root, so strokes inside each group are sorted by a topological sort afterward. A stroke $T_i$ precedes another stroke $T_j$ if one of the following conditions holds:
\begin{itemize}
	\item $T_i$ is on the left of $T_j$, where their projection to the $y$-axis(but not the $x$-axis) intersect;
	\item $T_i$ is on top of $T_j$, where their projection to the $x$-axis(but not the $y$-axis) intersect.
\end{itemize}
Further ambiguities are resolved by using the coordinates of the top left corner of the bounding boxes.

\begin{figure}
	\centering
	\includegraphics[width=\linewidth]{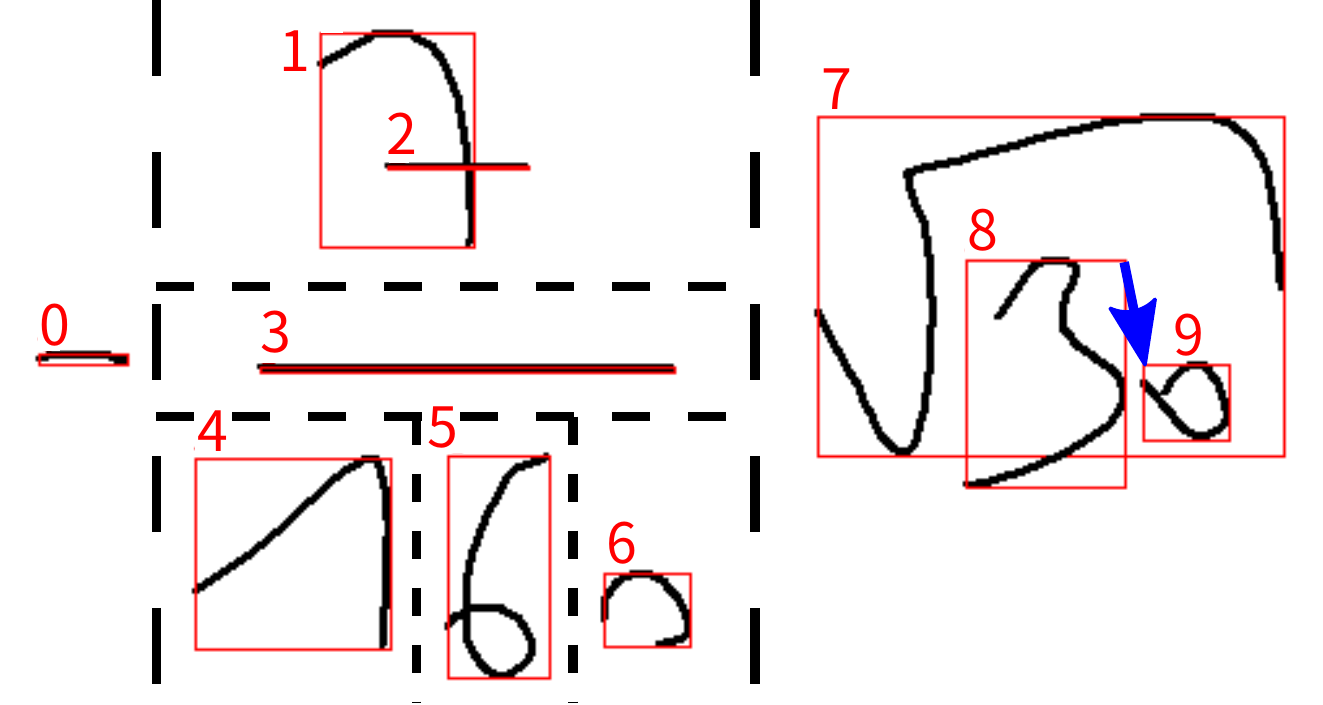}
	\caption{Stroke order normalization}\label{fig:ordering}
\end{figure}

Figure \ref{fig:ordering} illustrates how strokes are sorted, in which groups are separated by dotted lines and precedence relations are represented by arrows. 

\section{Evaluation}
\label{sec:evaluation}

\subsection{Datasets and evalation protocol}

The proposed system was evaluated on CROHME datasets, which are standard for mathematical expression recognition. Table \ref{tab:crohme} summarizes the datasets. For each mathematical expression in a dataset, list of points in each stroke is provided together with ground truth. In addition to MathML representation of the expression, correspondence between symbols and strokes is also provided. Since a bitmap image of mathematical expression can be obtained by rendering the strokes, the datasets can be used to evaluate offline recognition systems as well. Following the settings of task 2(offline handwritten formula recognition) in CROHME 2019, formulas were rendered at resolution of $1000\times 1000$ pixels by using the script provided by the organizers. However, it should be noted that rendered images are different from scanned or camera captured mathematical expressions in the level of background noises, although removing them is another research topic.

\begin{table}
	\caption{Number of expressions in CROHME datasets}
	\label{tab:crohme}
	\centering
	\begin{tabularx}{\linewidth}{Xrrr}
		\hline\noalign{\smallskip}
		Year&Training&Validation&Test\\
		2014&8836&671&986\\
		2016&8836&986&1147\\
		2019&9993&986&1199\\
		\noalign{\smallskip}\hline
	\end{tabularx}
\end{table}

Aligning with the offline task in CROHME 2019\cite{CROHME2019}, expression level metrics computed from the symbol level label graphs of formulas were used to evaluate the proposed system. Structure rate measures the percentage of recognized expressions matching the ground truth if all the labels of symbols are ignored. Expression rate measures the percentage of recognized expressions matching the ground truth up to a certain number of labeling mistakes on symbols or spatial relations. On the other hand, stroke classification rate, symbol segmentation rate, symbol recognition rate, and metrics based on stroke level label graph are inapplicable because offline recognition does not produce correspondence between symbols and strokes.

\subsection{Integration with a ready-made online recognizer}
\label{subsec:myscript}

In the first experiment, the proposed stroke extractor \footnote{The proposed stroke extractor is publicly available as a free software at \url{https://github.com/chungkwong/mathocr-myscript} and \url{https://github.com/chungkwong/mathocr-myscript-android}.} was combined with version 1.3 of MyScript Math recognizer\footnote{Documents of the recognizer is available at \url{https://developer.myscript.com/docs/interactive-ink/1.3/overview/about/}.}, which is a state-of-the-art online handwritten mathematical expression recognition system, to form an offline recognizer. The online recognizer was customized with a grammar to eliminate candidates of symbols and constructs that never appeared in CROHME datasets. Table \ref{tab:customizedgrammar} shows the customized grammar, where ``$<start>$'' is the start symbol.

\begin{table}
	\caption{Customized grammar for CROHME datasets}
	\label{tab:customizedgrammar}
	\centering
	\begin{tabularx}{\linewidth}{rrl}
		$<start>$ &::=& $<term>$\\
		$<term>$ &::=& $<operand>$\\
		&$|$& $<symbol2>$\\
		&$|$& $<symbol4>$\\
		&$|$& $<term> <term>$\\
		&$|$& $\frac{<term>}{<term>}$\\
		&$|$& $<operand>_{<term>}$\\
		&$|$& $<operand>^{<term>}$\\
		&$|$& $<operand>_{<term>}^{<term>}$\\
		&$\Big|$& $\begin{matrix}<symbol3>\\<term>\end{matrix}$\\
		&$\Bigg|$& $\begin{matrix}<term>\\<symbol4>\\<term>\end{matrix}$\\
		$<operand>$ &::=& $<symbol1>$\\
		&$|$& $\sqrt{<term>}$\\
		&$|$& $\sqrt[<term>]{<term>}$\\
		&$|$& $<symbol5> <term> <symbol6>$\\
		$<symbol1>$ &::=& `$0$' $|$ `$1$' $|$ `$2$' $|$ `$3$' $|$ `$4$' $|$ `$5$' $|$ `$6$' $|$ `$7$' $|$ `$8$' $|$ `$9$' \\
		&$|$& `$A$' $|$ `$B$' $|$ `$C$' $|$ `$E$' $|$ `$F$' $|$ `$G$' $|$ `$H$' $|$ `$I$' $|$ `$L$' \\
		&$|$& `$M$' $|$ `$N$' $|$ `$P$' $|$ `$R$' $|$ `$S$' $|$ `$T$' $|$ `$V$' $|$ `$X$' $|$ `$Y$' \\
		&$|$& `$a$' $|$ `$b$' $|$ `$c$' $|$ `$d$' $|$ `$e$' $|$ `$f$' $|$ `$g$' $|$ `$h$' $|$ `$i$' $|$ `$j$' $|$ `$k$' \\
		&$|$& `$l$' $|$ `$m$' $|$ `$n$' $|$ `$o$' $|$ `$p$' $|$ `$q$' $|$ `$r$' $|$ `$s$' $|$ `$t$' $|$ `$u$' \\
		&$|$& `$v$' $|$ `$w$' $|$ `$x$' $|$ `$y$' $|$ `$z$' $|$ `$\Delta$' $|$ `$\alpha$' $|$ `$\beta$' $|$ `$\gamma$' \\
		&$|$& `$\theta$' $|$ `$\lambda$' $|$ `$\pi$' $|$ `$\sigma$' $|$ `$\varphi$' $|$ `$\phi$' $|$ `$\mu$' $|$ `$<$' $|$ `$>$' $|$ `$!$' \\
		&$|$& `$/$' $|$ `$\infty$' $|$ `$\sin$' $|$ `$\cos$' $|$ `$\tan$' $|$ `$\lim$' $|$ `$\log$'\\
		$<symbol2>$ &::=& `$+$' $|$ `$-$' $|$ `$\pm$' $|$ `$\times$' $|$ `$\div$' $|$ `$\cdot$' $|$ `$=$' $|$ `$\prime$' $|$ `$,$' $|$ `$.$' \\
		&$|$& `$\ldots$' $|$ `$\rightarrow$' $|$ `$\exists$' $|$ `$\in$' $|$ `$\neq$' $|$ `$\leq$' $|$ `$\geq$' $|$ `$\forall$'\\
		$<symbol3>$ &::=& `$\sum$' $|$ `$\int$' $|$ `$\lim$'\\
		$<symbol4>$ &::=& `$\sum$' $|$ `$\int$'\\
		$<symbol5>$ &::=& `$($' $|$ `$[$' $|$ `$\{$' $|$ `$|$'\\
		$<symbol6>$ &::=& `$)$' $|$ `$]$' $|$ `$\}$' $|$ `$|$'
	\end{tabularx}
\end{table}

Table \ref{tab:crohme2014} shows experimental results on CROHME 2014 test set. The first seven are online recognition systems participated in CROHME 2014\cite{CROHME2014}. The proposed system outperformed all participated systems in CROHME 2014 except MyScript. Since MyScript Math recognizer itself has evolved over the past few years, MyScript Interactive Ink version 1.3, which is up-to-date as of this writing, was evaluated in the online setting too. On the other hand, im2markup
\cite{pmlr-v70-deng17a}, Watch, Attend, and Parse(WAP)\cite{ZHANG2017196}, CNN-BLSTM-LSTM\cite{LE2019255}, Paired Adversarial Learning(PAL)\cite{PAL}, and Multi-Scale Attention with Dense Encoder(MSD)\cite{8546031} are offline recognition systems, the proposed procedure achieved a better performance than them. Noted that the best results in the paper introducing WAP used online information to achieve an additional 2\% gain in expression rate.

\begin{table}
	\caption{Recognition performance on CROHME 2014 test set}
	\label{tab:crohme2014}
	\centering
	\begin{tabularx}{\linewidth}{Xrrrr}
		\hline\noalign{\smallskip}
		System & \multicolumn{3}{c}{Expression rate}  & Structure \\
		& Exact & $\leq 1$ & $\leq 2$&  rate \\
		& & label &  label &\\
		& & error & errors &\\
		\noalign{\smallskip}\hline\noalign{\smallskip}
		\multicolumn{5}{l}{Online recognizer}\\
		S\~{a}o Paulo&15.01&22.31&26.57& -\\
		RIT, CIS&18.97&26.37&30.83& -\\
		RIT, DRPL&18.97&28.19&32.35& -\\
		Tokyo&25.66&33.16&35.90& -\\
		Nantes&26.06&33.87&38.54& -\\
		Polit\`{e}cnica de Val\`{e}ncia&37.22&44.22&47.26& -\\
		MyScript&62.68&72.31&75.15& -\\
		MyScript 1.3&69.47&78.30&81.03& 82.86\\
		\noalign{\smallskip}\hline\noalign{\smallskip}
		\multicolumn{5}{l}{Offline recognizer}\\
		Harvard, im2markup&39.96&-&-&-\\
		USTC, WAP&44.4&58.4&62.2&-\\
		CAS, PAL&47.06&63.49&72.31&-\\
		TDTU, CNN-BLSTM-LSTM&48.78&63.39&70.18&-\\
		USTC, MSD&52.8&68.1&72.0&-\\
		\textbf{Proposed+MyScript 1.3}&58.22&71.60&75.15&77.38\\
		\noalign{\smallskip}\hline
	\end{tabularx}
\end{table}

Table \ref{tab:crohme2016} shows experimental results on CROHME 2016 test set. The first five are online recognition systems participated in CROHME 2016\cite{Mouchere2017ICFHR2016}. As expected, MyScript had a higher accuracy than the proposed system in all the metrics because the later was based on the former. The proposed system significantly outperformed all the remaining participated systems in CROHME 2016 without access to original strokes. On the other hand, WAP\cite{ZHANG2017196}, CNN-BLSTM-LSTM\cite{LE2019255}, and MSD\cite{8546031} are also offline recognition systems, the proposed system outperformed them.

\begin{table}
	\caption{Recognition performance on CROHME 2016 test set}
	\label{tab:crohme2016}
	\centering
	\begin{tabularx}{\linewidth}{Xrrrr}
		\hline\noalign{\smallskip}
		System & \multicolumn{3}{c}{Expression rate}  & Structure \\
		& Exact & $\leq 1$ & $\leq 2$&  rate \\
		& & label &  label &\\
		& & error & errors &\\
		\noalign{\smallskip}\hline\noalign{\smallskip}
		\multicolumn{5}{l}{Online recognizer}\\
		Nantes&13.34&21.02&28.33& 21.45\\
		S\~{a}o Paolo&33.39&43.50&49.17& 57.02\\
		Tokyo&43.94&50.91&53.70& 61.55\\
		Wiris&49.61&60.42&64.69& 74.28\\
		MyScript&67.65&75.59&79.86& 88.14\\
		MyScript 1.3&73.06&82.30&87.10& 88.58\\
		\noalign{\smallskip}\hline\noalign{\smallskip}
		\multicolumn{5}{l}{Offline recognizer}\\
		USTC, WAP&42.0&55.1&59.3&-\\
		TDTU, CNN-BLSTM-LSTM&45.60&59.29&65.65&-\\
		USTC, MSD&50.1&63.8&67.4&-\\
		\textbf{Proposed+MyScript 1.3}&65.65&77.68&82.56& 85.00\\
		\noalign{\smallskip}\hline
	\end{tabularx}
\end{table}

The proposed system participated in CROHME 2019\cite{CROHME2019}. Table \ref{tab:crohme2019} shows the final results of the competition. The proposed system was ranked the third place in the offline task.

\begin{table}
	\caption{Recognition performance on CROHME 2019 test set}
	\label{tab:crohme2019}
	\centering
	\begin{tabularx}{\linewidth}{Xrrrr}
		\hline\noalign{\smallskip}
		System & \multicolumn{3}{c}{Expression rate}  & Structure \\
		& Exact & $\leq 1$ & $\leq 2$&  rate \\
		& & label &  label &\\
		& & error & errors &\\
		\noalign{\smallskip}\hline\noalign{\smallskip}
		\multicolumn{5}{l}{Online recognizer}\\
		TUAT&39.95&52.21&56.54& 58.22\\
		MathType&60.13&74.40&78.57& 79.15\\
		PAL-v2&62.55&74.98&78.40& 79.15\\
		Samsung R\&D 2&65.97&77.81&81.73& 82.82\\
		MyScript 1.3&77.40&85.82&87.99& 88.82\\
		MyScript&79.15&86.82&89.82& 90.66\\
		Samsung R\&D 1&79.82&87.82&89.15& 89.32\\
		USTC-iFLYTEK&80.73&88.99&90.74&91.49\\
		\noalign{\smallskip}\hline\noalign{\smallskip}
		\multicolumn{5}{l}{Offline recognizer}\\
		TUAT&24.10&35.53&43.12&43.70\\
		Univ. Linz&41.49&54.13&58.88&60.02\\
		SCST-USTC&62.14&75.06&78.23&78.32\\
		PAL-v2& 62.89&74.98&78.40&79.32\\
		\textbf{Proposed+MyScript 1.3}& 65.22&78.48&83.07&84.90\\
		PAL& 71.23&80.31&82.65&83.82\\
		USTC-iFLYTEK& 77.15&86.82&88.99&89.49\\
		\noalign{\smallskip}\hline
	\end{tabularx}
\end{table}

Although the accuracy of the proposed offline recognizer was good compared with other offline recognizers, it was still much lower than that of the underlying online recognizer. If the stroke extractor worked perfectly, the two should be the same. The gap in expression rate was much larger than the gap in structure rate, the observation indicates that structural analysis is less sensitive to errors in stroke extraction than symbol recognition.

In order to narrow the gap, one should bring output of the stroke extractor and expected input of the online recognizer closer. There are two ways to achieve that. The first way is to improve the stroke extractor, so that it can recover written strokes exactly. The second way is to retrain the online recognition engine, so that it can adapt to the artificial strokes.

Improving the stroke extractor seems to be the most obvious choice, but it is hard. Adding more heuristic rules would lead to serious maintainability issues, whereas marginal benefit is diminishing. Therefore, many researchers had given up in this direction. Developing data driven stroke extractors is possible, but it would defeat the purpose. If heavy models such as convolutional and recurrent neural networks are applied, it is pointless to predict trajectories instead of formulas themselves.

Retraining the online recognition engine is a realistic choice. If the recognizer has seen artificial strokes given by the stroke extractor during training, it should be able to learn what they mean. By applying this technique, the sense of stroke extracted is decoupled from the way written by human beings, thus the stroke extractor needs not worry too much about corner cases. Unfortunately, MyScript Math recognizer was not trainable by user, so the next experiment was switched to another online recognizer.

\subsection{Integration with a trainable online recognizer}
\label{subsec:tap}

In the second experiment, the proposed stroke extractor was combined with TAP\footnote{The original version of TAP is available at \url{https://github.com/JianshuZhang/TAP}. Our version of programs, datasets and pretrained models are available at \url{https://github.com/chungkwong/mathocr-tap}.}, which is a published online handwritten mathematical expression recognition system\cite{8373726}, to form an offline recognizer. The paper introducing TAP combined three online models, three offline models, and three language models to get the best results, but ensemble modeling was not used in this experiment, since the objective is to check if a pure online model can adapt to the artificial strokes.

Evaluation was performed on the CROHME 2016 dataset, only the official training set and validation set were used to train and validate the models respectively. The encoder of TAP consumes an 8-dimensional feature vector for a point at each time step, the decoder of TAP predicts a TeX token at each time step, the loss function involves alignment between points and TeX tokens. When trained on written strokes, trace points and alignment were available from the dataset directly. When trained on artificial strokes, trace points were extracted from rendered images and alignment were estimated with Hausdorff distance. In both cases, TeX code for each formula was converted from annotation in MathML.

Table \ref{tab:crohme2016tap} shows experimental results on CROHME 2016 test set. The model which had been trained on written strokes performed much better on written strokes than on extracted strokes, the same phenomenon was already observed with MyScript. However, the model which had been trained on extracted strokes leaded to a much better offline recognizer, its accuracy almost caught up the online recognizer. The gap in expression rate between them was only 0.61\%, noted that the gap in expression rate between the best online system and the best offline system participated in CROHME 2019 was 3.58\%. The results verify that an online recognition system can adapted to the artificial strokes well by retraining. Another implication is that the essential difficulty between online recognition and offline recognition may be smaller than it was thought.

\begin{table}
	\caption{Recognition performance on CROHME 2016 test set given different training data}
	\label{tab:crohme2016tap}
	\centering
	\begin{tabularx}{\linewidth}{llrrrr}
		\hline\noalign{\smallskip}
		Training/validation& Test & \multicolumn{3}{c}{Expression rate}  & Structure \\
		&& Exact & $\leq 1$ & $\leq 2$&  rate \\
		&& & label &  label &\\
		&& & error & errors &\\
		\noalign{\smallskip}\hline\noalign{\smallskip}
		Written&Written&43.68&55.88&61.29&62.60\\
		Written&Extracted&23.63&38.71&47.17&51.79\\
		Extracted&Extracted&43.07&56.67&62.95&64.95\\
		\noalign{\smallskip}\hline
	\end{tabularx}
\end{table}


Although the offline recognizer built upon TAP was not as good as the one built upon MyScript, there is room for improvement. For example, language models, ensemble models, augmented datasets, and expanded datasets were not applied in this experiment, whereas they were commonly used by other state-of-the-art mathematical expression recognition systems\cite{CROHME2019} to boost the accuracy significantly\cite{8373726,LE2019255}.

\subsection{Efficiency}
\label{subsec:efficency} 

In the last experiment, efficiency of the proposed system was examined on devices ranging from low-end mobile phones to powerful GPU server. Table \ref{tab:device} shows details of those devices.

\begin{table}
	\caption{Specifications of tested devices}
	\label{tab:device}
	\centering
	\begin{tabularx}{\linewidth}{llX}
		\hline\noalign{\smallskip}
		Device&Component&Specification\\
		\noalign{\smallskip}\hline\noalign{\smallskip}
		Phone 1&CPU&Cortex-A7 4 cores @ 1.3 GHz \\
		&RAM&512MB\\
		Phone 2&CPU&Cortex-A53 2 cores @ 2.0 GHz + 6 cores @ 1.45 GHz \\
		&RAM&3GB\\
		Ultrabook&CPU&Intel\textregistered\ Core\texttrademark\ m3-6Y30 4 cores @ 0.90GHz\\
		&RAM&4GB\\
		Desktop&CPU&Intel\textregistered\ Core\texttrademark\ i5-7500 4 cores @ 3.40GHz\\
		&RAM&8GB\\
		Server&CPU&Intel\textregistered\ Xeon\textregistered\ Gold 6271C 2 cores @ 2.60GHz\\
		&RAM&32GB\\
		&GPU&NVIDIA\textregistered\ Tesla\textregistered\ V100 16GB\\
		\noalign{\smallskip}\hline
	\end{tabularx}
\end{table}

Table \ref{tab:time} shows the elapsed time used to recognize 1147 rendered mathematical expressions in the test set of CROHME 2016. Although the stroke extraction algorithm was implemented on CPU and not optimized for speed, it was still much faster than online recognition on all the devices tested. A modern low-end mobile phone(Phone 2) took about one second to recognize an image on average, whereas a clearly outdated mobile phone(Phone 1) took about two seconds. Although it is noticeable, the speed should be acceptable for on-device offline handwritten mathematical expression recognition, if users feel that online recognition is already fast enough.

\begin{table}
	\caption{Time used to recognize 1147 expressions in CROHME 2016 dataset}
	\label{tab:time}
	\centering
	\begin{tabularx}{\linewidth}{llrrrr}
		\hline\noalign{\smallskip}
		Device&Recognizer&\multicolumn{4}{c}{Execution time(s)}\\
		&&Stroke&Online&Total&Mean\\
		&&extraction&recognition&&\\
		\noalign{\smallskip}\hline\noalign{\smallskip}
		Phone 1&MyScript&1025&1653&2678&2.33\\
		Phone 2&MyScript&254&1123&1377&1.20\\
		Ultrabook&TAP&22&6908&6930&6.04\\
		Desktop&TAP&10&535&544&0.47\\
		&WAP&-&-&7845&6.84\\
		Server&TAP&34&152&186&0.16\\
		&WAP&-&-&417&0.36\\
		\noalign{\smallskip}\hline
	\end{tabularx}
\end{table}

WAP\footnote{WAP is avaliable at \url{https://github.com/JianshuZhang/WAP}.}, which is a native offline recognizer\cite{ZHANG2017196}, was slower than the proposed procedure on all the device tested. Like other modern optical character recognition systems, WAP uses a convolutional neural network which is computationally intensive\cite{Samsung} to extract features, so the speed relies heavily on the availability of powerful GPU. Since online recognition engines are usually less resource-hogging than native offline recognition engines, on-device online recognition is already integrated into various note-taking applications and input methods nowadays, whereas optical character recognition functionality usually relies on cloud based services. Therefore, stroke extraction provides an attractive way to improve user experience by avoiding unpredictable network latency and privacy issues.

\section{Conclusion}
\label{sec:conclusion}

In this paper, a stroke extraction algorithm is proposed for handwritten mathematical expression, so that an offline recognizer can be built upon an online one. A proof-of-concept implementation of the proposed stroke extractor is publicly available as a free software. Given a ready-made state-of-the-art online recognition engine, good offline accuracy was achieved on CROHME datasets. Given a trainable online recognition system, an offline recognizer with the same level of accuracy was constructed by retraining it with extracted strokes. Noted that the underlying online recognizers are not specially designed for extracted strokes.

The proposed approach is especially preferable for real-time use cases on devices with limited resources. Since online recognizers generally occupy less memory and run faster than native offline recognizers, stroke extraction provides an efficient way to implement on-device offline recognition on mobile phones and tablets.

Stroke extraction is a general methodology to offline handwriting recognition. Besides handwritten mathematical expression, the same approach can be applied to other types of handwriting such as chemical expression, musical notation, and diagram in principle. Examining if specialized stroke extractors are needed for different types of handwriting is a future work.

Developing independent recognition systems for online and offline handwriting may no longer be necessary, since stroke extraction allows advances on online recognition to be propagated immediately to the offline case. Instead, online recognition system makers can enter the offline market without abandoning existing investments. Therefore, the potential of reduction from offline recognition to online recognition is justified. 

\section*{Acknowledgments}

This research did not receive any specific grant from funding agencies in the public, commercial, or not-for-profit sectors.

\bibliographystyle{main} 
\bibliography{main}

\end{document}